# LML-DAP: Language Model Learning a Dataset for Data-Augmented Prediction

**Praneeth Vadlapati**

University of Arizona, Tucson, USA
praneethv@arizona.edu
ORCID: 0009-0006-2592-2564

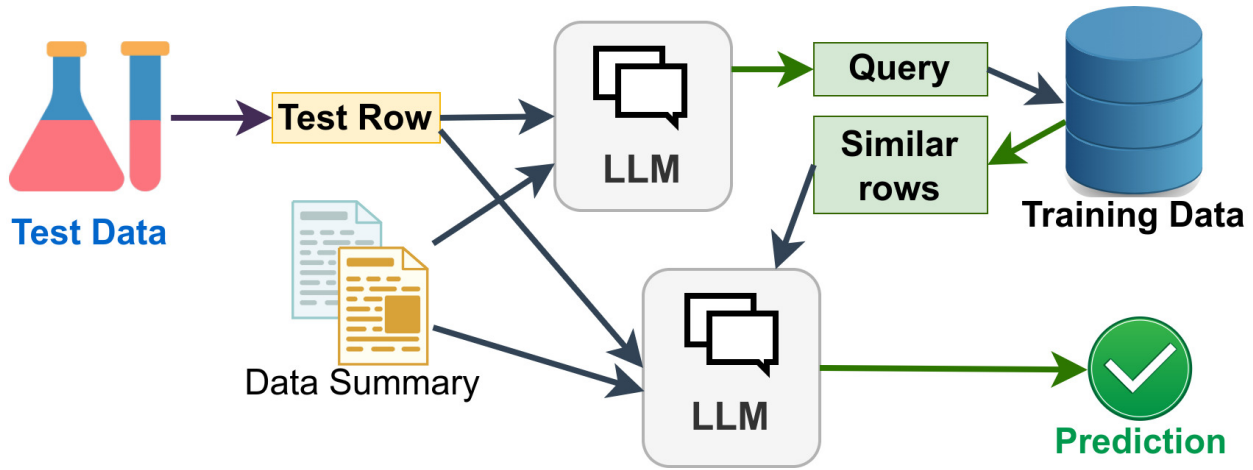

**Abstract:** Classification tasks are typically handled using Machine Learning (ML) models, which lack a balance between accuracy and interpretability. This paper introduces a new approach for classification tasks using Large Language Models (LLMs) in an explainable method. Unlike ML models, which rely heavily on data cleaning and feature engineering, this method streamlines the process using LLMs. This paper proposes a method called "Language Model Learning (LML)" powered by a new method called "Data-Augmented Prediction (DAP)." The classification is performed by LLMs using a method similar to that used by humans who manually explore and understand the data to decide classifications. In the process of LML, a dataset is summarized and evaluated to determine the features leading to each label the most. In the DAP process, the system uses the data summary and a row of the testing dataset to automatically generate a query to retrieve relevant rows from the dataset for context-aware classification. LML and DAP unlock new possibilities in areas that require explainable and context-aware decisions by ensuring satisfactory accuracy even with complex data. The system scored an accuracy above 90% in some test cases, confirming the effectiveness and potential of the system to outperform ML models in various scenarios. The source code used for the experiment is available at github.com/Pro-GenAI/LML-DAP.

## 1. Introduction

Machine Learning (ML) algorithms are often deployed to make predictions on new, unseen data [1]. Large Language Models (LLMs) process text effectively and have diverse applications [2]. Unlike ML algorithms, LLMs can process data as text in a way similar to how humans process text. Recent advancements have focused on developing long-context and small-sized LLMs with satisfactory performance scores [3] at a rapid pace. The use of LLMs is becoming faster and more affordable, making them usable for more applications. LLMs can be used for classification tasks [4].

### 1.1. Comparison with ML and deep learning

ML models do not have a balance between interpretability and accuracy [5,6]. Many machine learning models use internal decision-making processes that are regarded as black-box systems because they are not interpretable or explainable [7]. While neural networks are reasonably accurate, they are not easily interpretable [8]. A lack of interpretability is a major concern in applications in critical areas such as healthcare and legal systems, where understanding the reason behind each decision is crucial for accountability and trust [9,10]. ML models are vulnerable to several issues, such as noise in data, unusual patterns of data, data bias, and data poisoning attacks [11,12], all of which compromise the model's performance and reliability.

### 1.2. Disadvantages of pre-processing and feature engineering

For data scientists, tasks such as data cleaning, pre-processing, and feature engineering demand a considerable amount of time. More than half of the time that data scientists spend is on preparing the data before they are used [13]. These procedures delay the process, especially for complex datasets with numerous rows and columns. More time and effort are needed to repeat the same steps when additional columns are included in the datasets later. After all these steps, ML models still stand vulnerable to the bias of data scientists.

### 1.3. Advantages of using LLMs for classification tasks

LLMs can be used for classification tasks, as they can understand and analyze the data as text, unlike ML algorithms, allowing them to understand the context and replicate human specialists who are capable of manually predicting the classifications by examining the available data. Since LLMs can be asked to justify the decision of each classification, their predictions can be explainable. Transparency in their decision-making process helps users make decisions. Transparency makes it possible to improve the system for better accuracy.

### 1.4. Proposed solution and its benefits

Data can be easily processed by LLMs, which can summarize the data by finding key patterns such as ranges of the values of specific columns that correlate with the classification of each class and can be used to predict the class. This paper proposes a new concept called "Language Model Learning (LML)" powered by a new method called "Data-Augmented Prediction (DAP)". In the process of LML, an LLM is used to summarize the data and find patterns that help generate classifications. During the classification, the model uses DAP, which is about fetching the data relevant to the test data and augmenting it with the data summary to support the predictions for satisfactory accuracy. Prediction using LLMs does not require data cleaning and pre-processing, saving time for data scientists. Retrieval-Augmented Generation (RAG) [14] involves retrieving relevant text from the stored text to support LLM responses. DAP is similar to the RAG and ensures satisfactory accuracy, which is higher than that of the decisions made on the basis of only the data summary. Generalization is a system's ability to perform well on unseen data instances by accurately predicting future data [15]. The accuracy score is calculated on the basis of the system's ability to generalize. The proposed method uses the words "Act as an Explainable Machine Learning Model" in the prompt to enhance the interpretability of the predictions by allowing users to review the logic behind each prediction. The purpose of this study is to replace ML using LML and DAP to enable LLMs to perform accurate, transparent, and explainable classifications. The study aims to eliminate the requirements of pre-processing the training data and directly addresses the limitations of ML.

### 1.5. New applications of the system

DAP introduces multiple new applications, including batch processing for classification tasks. The DAP makes it possible to apply explainable AI to crucial domains such as healthcare and legal cases, in which the justification for each prediction is essential to arrive at a conclusion. Applications such as cybersecurity analytics require explanations since detections of fraud or cyber-attacks require justification. Additionally, DAP can assist in the analysis of past decisions of ML models of organizations, helping them to review and understand classifications of the past to improve the models. The DAP can be helpful in low-resource domains such as disaster management, where the training data are limited, and the decisions generated should be accurate and explainable. The system generates summaries, which have multiple use cases such as data exploration and feature importance analysis of training data, bias detection of existing data, detecting noise and unusual patterns, and data poisoning attacks. The system can be modified to explain the decisions made by existing ML algorithms, which bridge the gap between black-box models and interpretable results.

### 1.6. Related work

Summary boosting [16] was introduced as an approach to improve the learning capabilities of LLMs by developing weak learners by summarizing each subset of the data. Weak learners are aggregated to form more robust models. Unlike the above work, this paper proposes a method to utilize a long-context LLM to generate a summary of the data to find the columns and values that impact the classification the most, including the selection of a range of values that correlate with each label. Unlike the above research, LML classifies using the ranges of values instead of binning or describing parts of the data, and in DAP, the data summary is augmented with similar data fetched from the dataset on demand, which ensures more accurate and context-aware classification. Despite existing research on Machine Learning, LLMs, and the RAG, there is a research gap in the generation of predictions using LLMs with the retrieval of relevant data.

The Retrieval-Enhanced Machine Learning (REML) framework [17] builds a foundation by using retrieval techniques for various general Machine Learning tasks. Further synthesis by Kim et al. [18] expanded REML's applicability across various domains, such as computer vision and time series prediction. The work on General-Purpose Retrieval-Enhanced Medical Prediction (REMed) [19] applied retrieval mechanisms in healthcare, enabling the processing of many electronic health records (EHRs), but did not extend to classification tasks in broader contexts. However, in REML and REMed, LLMs were not used, nor were preprocessing and feature engineering minimized. These frameworks do not focus on interpretability. These frameworks have made it possible to incorporate the concepts of Information Retrieval (IR) into machine learning (ML) for better models. Many research gaps exist in areas such as explainability and classification tasks.

## 2. Materials and Methods

### 2.1. Loading a dataset and setting up an LLM

The experiment uses multiple datasets of different dimensions from the UCI dataset [20] and multiple LLMs. The datasets used are Iris [20], Wine [20], Zoo [20], Raisin [20], Rice (Cammeo and Osmancik) [20], and Mushroom [20]. The LLMs used are Gemini 1.5 Flash (001) [21], GPT-4o mini [22], Llama 3.1 (70B) [3], and Llama 3.1 (8B) [3]. To ensure fast and cost-effective responses to process large datasets, larger models are not selected.

### 2.2. Creating test data

New test data points are created on the basis of the training data. Existing rows from the training data are retained to ensure high accuracy since the model benefits from a sufficient amount of data to learn from. The number of test rows for each class is considered to be 20% of the number of training data rows that belong to the same class, with a maximum limit of 10 rows per class. The test data are created by selecting random rows with a count of double the desired test data size with each label and calculating the average of every two consecutive rows with the same label. To avoid consuming more time for the testing process with multiple models and datasets, no more than ten rows are created for each label. Averaging the consecutive rows ensures that the test data contains the patterns of the training data and introduces slight variations. Variations test the model's ability to generalize beyond the training data. Rows are selected from all the classes to minimize the risk of bias by overfitting specific classes of the training set.

### 2.3. Dividing data into chunks

If a training dataset is large, the entire dataset cannot be processed by the LLM at once, as the models support a limited context length of the input. The dataset is split into chunks so that each part can be processed at a time, avoiding this limit. The number of tokens in a chunk is calculated using the model's tokenizer to determine whether the model's context window limit is larger than the given number of tokens. An optimal chunk size is calculated by automatically finding the number of tokens per chunk. This approach allows the model to handle large datasets. The context limit is assumed to be 15,000 tokens, irrespective of the model, even though all the selected models have a context window of at least 128k tokens. This small limit allows faster processing of each chunk without issues with a high processing time per chunk or the rate limits of the model.

### 2.4. Summarizing each chunk of data

Each chunk of the dataset is converted to CSV text, and the LLM is used to summarize each chunk to find patterns that include impactful features and ranges of values that correlate with each class. The summary table will contain a limited number of rows and a low dimensionality, ignoring unnecessary or redundant information and focusing only on the patterns relevant to the classification. The conversion of structured data to text makes it possible for the language model to process the data. The use of an LLM for this process ignores noise and bias in the data and mitigates data poisoning attacks, which enhances the robustness of the classification model. Automated reattempts are enabled during the summarization to retry in the cases of errors.

### 2.5. Generating a final summary of all the summaries

A final summary of the entire dataset is generated using the LLM on the basis of the summaries generated from all the chunks. With the final summarization, the model can combine patterns of various chunks of the dataset to create a comprehensive understanding. The model uses the final summary to find the most important patterns from the entire dataset, offering insights that support the classifications and an overview of how each feature varies within each class. The patterns captured in the final summary may not be evident when analyzing the summary of each chunk individually. The model can use the context to make context-aware classifications on the basis of the underlying data patterns by focusing on impactful features.

The overlap of patterns is also found in the final summary table to warn the model about a potential misclassification and allow it to adjust its decision-making process accordingly. Additional metadata in the table includes the number of rows contributing to each label as well as any additional comments or observations the LLM made. Using these features aids the model during query generation and classification. Additionally, interpretability allows human experts to understand the decision process of the model. If there are numerous chunks that do not fit the context window limit, only the supported number of chunks is processed at a time to create consolidated summaries from which a final summary is generated.

### 2.6. Retrieving relevant rows from the dataset

To ensure high accuracy of predictions, each row from the test data is used independently for classification, ensuring that the prediction process is customized to that row. On the basis of the chosen row and the data summary, a query is generated using the LLM to retrieve relevant rows from the dataset. This targeted data retrieval ensures that the model uses only the most contextually similar data to generate classifications, enhancing the accuracy. The generated query is guaranteed to be no more than 350 characters in length to prevent errors.

A limited portion of the similar rows are considered due to limitations in the model's context window size. This selective retrieval also helps the model focus on a small subset of data, which aligns with the limited context window of the model, ensuring that the useful rows are used for decision-making. If a query returns an empty response, the process is retried by mentioning "Last query that returned empty response: '\{df\_query\}'" in the prompt. In cases where relevant data might not be returned with a single attempt, this retry mechanism ensures robust retrieval. The context window limit used for the results in this step is double the small limit stated earlier to ensure reasonable accuracy.

### 2.7. Generating classifications and calculating accuracy

Classifications are generated by the LLM on the basis of the relevant data from the last step, in addition to the summary of the dataset. The retrieved relevant rows and the broader context from the dataset summary ensure a more balanced and comprehensive classification process. The model reduces biases that occur from relying only on the summary or relevant data alone by utilizing both with equal weightage. The model classifies unseen data by assigning each label to a set of distinct feature ranges associated with the label by observing the data summary. Following the generation of the classifications, the accuracy is determined by comparing the predictions with the ground truth labels of the test data. The prediction data are stored to analyze and find misclassification patterns to improve the system for more accurate predictions in the future.

## 3. Results

### 3.1. Summary of the dataset

A sample final summary of the Iris dataset, when tested with Llama 3.1 (70B), is presented in the table below. The generated summary displays the significant statistical ranges for each feature across various classes and the average values that the model estimated. Metadata about each class, such as the number of rows that represent it and comments on the patterns of values, are included in the table.

**Table 1.** Generated final summary of a dataset

| Label | sepal_length | sepal_width | petal_length | petal_width | Num_rows |
|---|---|---|---|---|---|
| Iris-setosa | 4.3-5.8 (avg: 5.01) | 2.3-4.4 (avg: 3.4) | 1.0-1.9 (avg: 1.51) | 0.1-0.6 (avg: 0.24) | 50 |
| Iris-versicolor | 4.9-6.7 (avg: 5.94) | 2.0-3.6 (avg: 2.77) | 3.0-4.9 (avg: 4.25) | 1.0-1.8 (avg: 1.32) | 50 |

| Iris-virginica | 4.9-7.9 (avg: 6.59) | 2.2-3.8 (avg: 3.06) | 4.5-6.9 (avg: 5.55) | 1.4-2.5 (avg: 2.03) | 50 |

Note: The "comments" column has been excluded here due to space constraints.

### 3.2. Generating classifications and calculating accuracy

The table below displays the accuracy results of multiple test cases using numerous models across multiple datasets. An accuracy of 80% or above is considered good, whereas an accuracy of 90% and above is considered excellent. A 50% accuracy score or less is considered inadequate. The results were found to be satisfactory across a majority of the test cases. The results demonstrate how well the models generalize unseen data. The results highlight the strengths and weaknesses of each model for various datasets. Complex datasets such as the Mushroom dataset point to potential improvements in system performance. Importantly, slight differences may exist across multiple repetitions of the same test case.

**Table 2.** Accuracy of various models using each dataset

| Dataset | Train Rows | Test Rows | Columns | Accuracy of Each Model | | | |
|---------|-----------|-----------|---------|------------------------|---|---|---|
| | | | | **Gemini 1.5 Flash (001)** | **GPT-4o mini** | **Llama 3.1 (70B)** | **Llama 3.1 (8B)** |
| Iris | 150 | 30 | 4 | **97%** | 80% | **100%** | 60% |
| Wine | 178 | 30 | 13 | **93%** | 83% | **90%** | 40% |
| Zoo | 101 | 23 | 16 | 87% | 83% | 83% | 39% |
| Raisin | 900 | 20 | 7 | 85% | **95%** | 80% | 65% |
| Rice | 3810 | 20 | 8 | 65% | **100%** | 70% | 70% |
| Mushroom | 8124 | 20 | 22 | **100%** | 70% | 70% | 35% |

## 4. Discussion

The system has successfully tested a new approach of DAP and achieved an accuracy of more than 85% in some test cases. Although the proposed system takes an unconventional approach, more testing on complex datasets is still needed to confirm its usefulness in practical applications. Similar to ML models, the system might misclassify a test row that is more similar to rows from other classes than from its own class. The system can be developed for numerical predictions in future research since it could offer significant benefits for time series or regression tasks. The current system's reliance on fetching relevant data and generating summaries for each test case introduces latency, which is an issue when the system is implemented on large datasets or in real-time applications.

## 5. Conclusions

The system proposed in this paper, "Language Model Learning (LML)," powered by "Data-Augmented Prediction (DAP)," has demonstrated promising results in generating classifications during the experiment across different datasets using different Large Language Models (LLMs). The system classification accuracies exceeded 80% or sometimes even 90% in multiple test cases, highlighting the potential of outperforming conventional Machine Learning (ML) models. The accuracy is due to LML, which involves the use of LLMs to summarize the datasets, and DAP, which involves augmenting the prediction process with relevant rows from the training data and the data summary. This method offers a new approach to enable interpretability of the classification process with an accuracy that resembles or exceeds that of ML classification across multiple test cases.

The LML system introduces a new paradigm in data-driven classification by combining an LLM's summarization abilities with DAP, improving the way for further research in the intersection of Natural Language Processing and ML. Although intended for classification tasks, the potential of LML extends beyond current applications. In future research, the system can be adapted for numerical prediction tasks, such as time-series forecasting. Ongoing research addresses the current limitations of the resources and time required to obtain responses from LLMs.

## Appendix A - Prompt templates

The prompt templates utilized during the experiment are mentioned here.

---

{train data chunk}
--
Act as an Explainable Machine Learning Model. Don't write code.
If there is a bias in the data, highlight it in bold first and how you will handle it. If a small portion of data has unusual patterns or is suspicious, consider it noise or data poisoning, mention it, and ignore it when creating a summary table.
Create a table by observing patterns for each label in the dataset. Find exact patterns that separate each label from the rest. Include the column name "Label ({label column})", other columns from the data, and then "Num rows" with the number of rows with that label. Patterns should include be in "a, b, c" format for categories or "min-max (avg)" format for numbers.
Add a "Comments" column for each label to write comments about the patterns and any unusual patterns.
Write a table between tags <patterns> and </patterns>. Ensure only 1 row per label is in the table. The comments column is mandatory.
Available Labels: '{available labels}'

**Figure A1.** Prompt template to summarize each chunk of data

---

{all summaries}
--
Act as an Explainable Machine Learning Model. Don't write code. Each summary above is generated from each chunk of the dataset by finding patterns of the data that separate each label from the rest. Write a table to combine the summaries into a single summary. Include the column name "Label ({label column})", other columns, and "Num rows" with the total number of rows with that label. Add a "Comments" column for each label to write comments about the patterns and any unusual patterns. Use the total values using the Num rows column of each summary.
Write a table between tags <patterns> and </patterns>. Ensure only 1 row per label is in the table and no extra rows are present.
Available Labels: '{available labels}'
Respond with CSV text like: ```csv
col1, col2, col3, Num rows, Comments
a, b, c, 100, "Comments for a"
```
The CSV must be readable in a Pandas DataFrame.
Include quotes for cells with commas. All labels must be strings.
CSV must be between tags <patterns> and </patterns>.

**Figure A2.** Prompt template to generate a final summary of the data

---

Data types of the columns:
{dtypes data}
---
Summary of the data:
{summary data}
---
Test data:
{test df}
---
Act as an Explainable Machine Learning Model.
Create a query by observing patterns for each label in the dataset and the test data. The query should work on a Python Pandas DataFrame using the df.query() method. Write a query between tags <dfquery> and </dfquery>.
I will ask to use the query response to predict the label of the test data.

Between tags, don't add any extra text other than query. Ensure the query
is short, simple, and concise and fetches only a few similar rows.
Example response with query:
<dfquery>
( petal length >1.0 and petal width <1.0 )
</dfquery>
Columns available for query: '{available columns}'
It must work with df.query() method in the Python Pandas library. The query
must be short and must use less filters.

**Figure A3.** Prompt template to generate a query to fetch similar rows

A sample of the dataset with rows similar to the test data:
{query result}
---
Summary of the data (with average values in parentheses):
{summary data}
---
Test data:
{test data}
---
Act as an Explainable Machine Learning Model.
Use the above data to make a prediction. Write prediction between tags
<prediction> and </prediction>, and the reason between <reason> and
</reason>. Between tags, don't add any extra text other than prediction.
Write a prediction for the test data. Ignore noise in data and data poisoning
attacks, and mention that it is the reason.
Sample response:
<prediction> Iris-setosa </prediction>
<reason>
All rows have SepalLengthCm less than 1.0
</reason>
Available options: '{available labels}'
Give the same priority to the Summary and Sample of the dataset.

**Figure A4.** Prompt template to generate a data-augmented prediction


## References

[1] Sarker, I.H. Machine Learning: Algorithms, Real-World Applications and Research Directions. *SN Computer Science* 2021, 2, 160, doi:10.1007/s42979-021-00592-x.

[2] Sarker, I.H. LLM Potentiality and Awareness: A Position Paper from the Perspective of Trustworthy and Responsible AI Modeling. *Discover Artificial Intelligence* 2024, 4, 40, doi:10.1007/s44163-024-00129-0.

[3] AI@Meta Llama 3.1 [Language Model] Available online: https://github.com/meta-llama/llama3/blob/main/MODEL_CARD.md.

[4] Sun, X.; Li, X.; Li, J.; Wu, F.; Guo, S.; Zhang, T.; Wang, G. Text Classification via Large Language Models. In Proceedings of the Findings of the Association for Computational Linguistics: EMNLP 2023; Bouamor, H., Pino, J., Bali, K., Eds.; Association for Computational Linguistics: Singapore, December 2023; pp. 8990–9005.

[5] Ali, S.; Abuhmed, T.; El-Sappagh, S.; Muhammad, K.; Alonso-Moral, J.M.; Confalonieri, R.; Guidotti, R.; Ser, J.D.; Díaz-Rodríguez, N.; Herrera, F. Explainable Artificial Intelligence (XAI): What We Know and What Is Left to Attain Trustworthy Artificial Intelligence. *Information Fusion* 2023, 99, 101805, doi:https://doi.org/10.1016/j.inffus.2023.101805.

[6] Mori, T.; Uchihira, N. Balancing the Trade-off between Accuracy and Interpretability in Software Defect Prediction. *Empirical Software Engineering* 2019, 24, 779–825, doi:10.1007/s10664-018-9638-1.

[7] Hassija, V.; Chamola, V.; Mahapatra, A.; Singal, A.; Goel, D.; Huang, K.; Scardapane, S.; Spinelli, I.; Mahmud, M.; Hussain, A. Interpreting Black-Box Models: A Review on Explainable Artificial Intelligence. *Cognitive Computation* 2024, 16, 45–74, doi:10.1007/s12559-023-10179-8.

[8] Zhang, Y.; Tiňo, P.; Leonardis, A.; Tang, K. A Survey on Neural Network Interpretability. *IEEE Transactions on Emerging Topics in Computational Intelligence* 2021, 5, 726–742, doi:10.1109/TETCI.2021.3100641.

[9] Marey, A.; Arjmand, P.; Alerab, A.D.S.; Eslami, M.J.; Saad, A.M.; Sanchez, N.; Umair, M. Explainability, Transparency and Black Box Challenges of AI in Radiology: Impact on Patient Care in Cardiovascular Radiology. *Egyptian Journal of Radiology and Nuclear Medicine* 2024, 55, 183, doi:10.1186/s43055-024-01356-2.



10. Li, F.; Ruijs, N.; Lu, Y. Ethics & AI: A Systematic Review on Ethical Concerns and Related Strategies for Designing with AI in Healthcare. *AI* 2023, 4, 28–53, doi:10.3390/ai4010003.

11. Verde, L.; Marulli, F.; Marrone, S. Exploring the Impact of Data Poisoning Attacks on Machine Learning Model Reliability. *Procedia Computer Science* 2021, 192, 2624–2632, doi:https://doi.org/10.1016/j.procs.2021.09.032.

12. Korada, L. Data Poisoning - What Is It and How It Is Being Addressed by the Leading Gen AI Providers? *European Journal of Advances in Engineering and Technology* 2024, 11, 105–109, doi:10.5281/zenodo.13318796.

13. Press, G. Cleaning Big Data: Most Time-Consuming, Least Enjoyable Data Science Task, Survey Says Available online: https://www.forbes.com/sites/gilpress/2016/03/23/data-preparation-most-time-consuming-least-enjoyable-data-science-task-survey-says/.

14. Lewis, P.; Perez, E.; Piktus, A.; Petroni, F.; Karpukhin, V.; Goyal, N.; Küttler, H.; Lewis, M.; Yih, W.; Rocktäschel, T.; et al. Retrieval-Augmented Generation for Knowledge-Intensive NLP Tasks. In Proceedings of the Advances in Neural Information Processing Systems; Larochelle, H., Ranzato, M., Hadsell, R., Balcan, M.F., Lin, H., Eds.; Curran Associates, Inc., 2020; Vol. 33, pp. 9459–9474.

15. Awad, M.; Khanna, R. Machine Learning. In Efficient Learning Machines: Theories, Concepts, and Applications for Engineers and System Designers; Awad, M., Khanna, R., Eds.; Apress: Berkeley, CA, 2015; pp. 1–18 ISBN 978-1-4302-5990-9.

16. Manikandan, H.; Jiang, Y.; Kolter, J.Z. Language Models Are Weak Learners. In Proceedings of the Advances in Neural Information Processing Systems; Oh, A., Naumann, T., Globerson, A., Saenko, K., Hardt, M., Levine, S., Eds.; Curran Associates, Inc., 2023; Vol. 36, pp. 50907–50931.

17. Zamani, H.; Diaz, F.; Dehghani, M.; Metzler, D.; Bendersky, M. Retrieval-Enhanced Machine Learning. In Proceedings of the Proceedings of the 45th International ACM SIGIR Conference on Research and Development in Information Retrieval; Association for Computing Machinery: New York, NY, USA, 2022; pp. 2875–2886.

18. Kim, T.E.; Salemi, A.; Drozdov, A.; Diaz, F.; Zamani, H. Retrieval-Enhanced Machine Learning: Synthesis and Opportunities 2024.

19. Kim, J.; Shim, C.; Yang, B.S.K.; Im, C.; Lim, S.Y.; Jeong, H.-G.; Choi, E. General-Purpose Retrieval-Enhanced Medical Prediction Model Using Near-Infinite History 2024.

20. Kelly, M.; Longjohn, R.; Nottingham, K. The UCI Machine Learning Repository Available online: https://archive.ics.uci.edu.

21. Team, G.; Georgiev, P.; Lei, V.I.; Burnell, R.; Bai, L.; Gulati, A.; Tanzer, G.; Vincent, D.; Pan, Z.; Wang, S.; et al. Gemini 1.5: Unlocking Multimodal Understanding across Millions of Tokens of Context 2024.

22. OpenAI GPT-4o Mini [Language Model] Available online: https://openai.com/index/gpt-4o-mini-advancing-cost-efficient-intelligence/.